\pdfoutput=1

\documentclass[11pt]{article}

\usepackage[final]{acl}

\usepackage{times}
\usepackage{latexsym}

\usepackage[T1]{fontenc}

\usepackage[utf8]{inputenc}

\usepackage{microtype}

\usepackage{inconsolata}

\usepackage{graphicx}

\usepackage{amsmath,amsfonts,bm}

\def\eqref#1{equation~\ref{#1}}

\def\1{\bm{1}}

\DeclareMathAlphabet{\mathsfit}{\encodingdefault}{\sfdefault}{m}{sl}
\SetMathAlphabet{\mathsfit}{bold}{\encodingdefault}{\sfdefault}{bx}{n}

\usepackage{algorithm}
\usepackage{algorithmic}
\usepackage{setspace}
\usepackage{comment}
\usepackage{booktabs}
\usepackage{hyperref}
\usepackage{icomma}
\usepackage{url}

\usepackage[nameinlink]{cleveref}

\usepackage{longtable}
\usepackage{xcolor}
\usepackage{colortbl}
\usepackage{multicol}
\usepackage{multirow}
\usepackage{makecell}
\usepackage{amsmath, amssymb}
\usepackage{mathtools}
\usepackage{enumitem}
\usepackage{subcaption}
\usepackage{float}
\usepackage{graphicx}
\usepackage{caption} 
\usepackage{upgreek}
\usepackage{seqsplit}
\usepackage{color,soul}
\usepackage{arydshln}
\usepackage{xspace}
\usepackage{adjustbox}
\usepackage{enumitem}
\usepackage{booktabs} 
\usepackage[most]{tcolorbox}

\definecolor{Red}{rgb}{0.6,0,0}
\definecolor{Blue}{rgb}{0,0,0.8}
\definecolor{Green}{rgb}{0.0,0.55,0.0}
\definecolor{mountainmeadow}{rgb}{0.19, 0.73, 0.56}
\definecolor{crimson}{rgb}{0.86, 0.08, 0.24}
\definecolor{darkblue}{rgb}{0.0, 0.0, 0.55}
\definecolor{ggr}{gray}{0.92}

\definecolor{gg}{HTML}{E0FEFE}

\newcolumntype{a}{>{\columncolor{ggr}}c}

\hypersetup{
    colorlinks = true,
    citecolor = darkblue,
    linkcolor = crimson,
    linktoc = all,
    urlcolor = darkblue,
}

\newcommand{\highlight}[1]{{\color{crimson}{#1}}}

\newcommand{\ours}{TAMP\xspace}
\newcommand{\relp}{Rel.\xspace} 
\newcommand{\jwl}[1]{{\color{black}{#1}}}

\NewDocumentCommand{\may}
{ mO{} }{\textcolor{blue}{\textsuperscript{\textit{May}}\textsf{\textbf{\small[#1]}}}}

\NewDocumentCommand{\san}
{ mO{} }{\textcolor{blue}{\textsuperscript{\textit{sandeep}}\textsf{\textbf{\small[#1]}}}}
\title{TAMP: Token-Adaptive Layerwise Pruning in \\Multimodal Large Language Models}

\author{
    Jaewoo Lee$^{1}$ \; 
    Keyang Xuan$^{2}$ \; 
    Chanakya Ekbote$^{3}$\\
    \textbf{Sandeep Polisetty}$^{4}$ \;
    \textbf{Yi R. (May) Fung}$^{5}$ \; 
    \textbf{Paul Pu Liang}$^{3}$ \\
    University of North Carolina Chapel Hill$^{1}$, University of Illinois Urbana-Champaign$^{2}$,\\ Massachusetts Institute of Technology$^{3}$, University of Massachusetts Amherst$^{4}$,\\ Hong Kong University of Science and Technology$^{5}$ \\
    \texttt{jwoolee@cs.unc.edu keyangx3@illinois.edu cekbote@mit.edu}\\
    \texttt{spolisetty@umass.edu yrfung@cse.ust.hk ppliang@mit.edu}
}

\begin{document}
\maketitle
\begin{abstract}
Multimodal Large Language Models (MLLMs) have shown remarkable versatility in understanding diverse multimodal data and tasks. However, these capabilities come with an increased model scale. While post-training pruning reduces model size in unimodal models, its application to MLLMs often yields limited success. Our analysis discovers that conventional methods fail to account for the unique token attributes across layers and modalities inherent to MLLMs. Inspired by this observation, we propose TAMP, a simple yet effective pruning framework tailored for MLLMs, featuring two key components: \textit{(1) Diversity-Aware Sparsity}, which adjusts sparsity ratio per layer based on diversities among multimodal output tokens, preserving more parameters in high-diversity layers; and \textit{(2) Adaptive Multimodal Input Activation}, which identifies representative multimodal input tokens using attention scores to guide unstructured weight pruning. We validate our method on two state-of-the-art MLLMs: LLaVA-NeXT, designed for vision-language tasks, and VideoLLaMA2, capable of processing audio, visual, and language modalities. Empirical experiments across various multimodal evaluation benchmarks demonstrate that each component of our approach substantially outperforms existing pruning techniques. Our code is available at \href{https://github.com/G-JWLee/TAMP}{\textcolor{magenta}{https://github.com/G-JWLee/TAMP}}.
\end{abstract}
\section{Introduction}
Large Language Models (LLMs) have achieved remarkable success at billion-parameter scales ~\citep{llama,llama2,deepseek}, excelling in challenging tasks. Building on this, Multimodal Large Language Models (MLLMs)~\citep{Li2024llavanext,Zhan2024anygpt,Wu2024nextgpt}, which extend LLMs to handle diverse modality inputs, have grown in size to address the complexities of multimodal tasks~\citep{liang2024foundations,Tong2024cambrian,Shi2024mathllava}. While beneficial for performance, their colossal model size imposes substantial computational and memory resources, limiting their practicality in resource-constrained scenarios~\citep{Reid2024gemini15pro, Li2024llavaonevision}.
\begin{figure}[t]
    \centering
    \includegraphics[width=\linewidth]{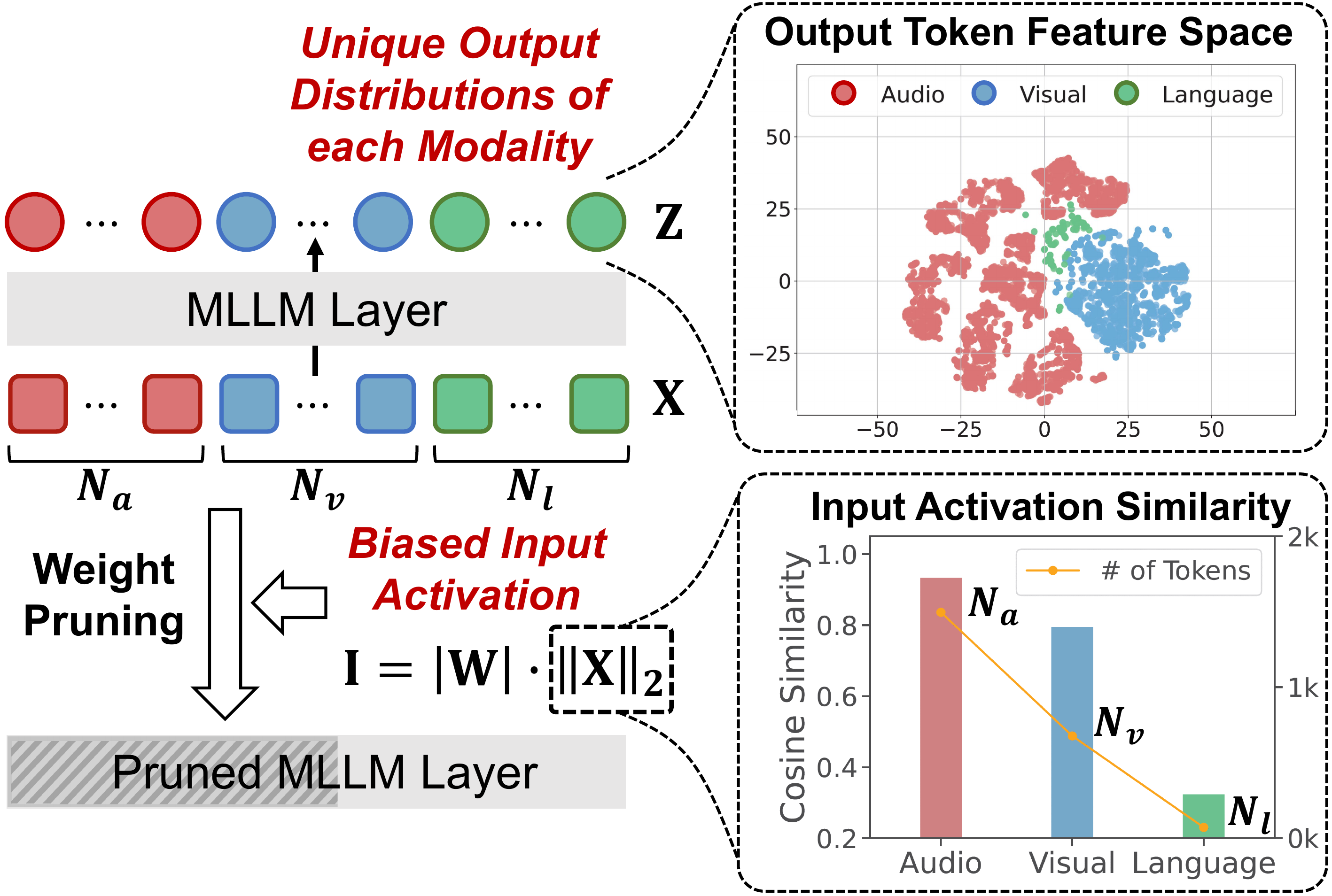}
    \par
    \caption{\small Illustration of multimodal token attributes. \textbf{(Top)}: t-SNE visualization of multimodal output tokens of the layer, exhibiting unique distributions of each modality. \textbf{(Bottom)}: Cosine similarity between the $\ell_{2}$-norm of tokens from each modality and all tokens, demonstrating a bias in input activations toward the modality with the largest token count ($N_a,N_v\!>>\!N_l$), resulting in suboptimal weight pruning.
    }
    \label{fig:multimodal_tok_one_col}
\end{figure}

Post-training model pruning~\citep{Sung2024wanda,Frantar2023sparsegpt,Ma2023llmpruner,Yu2023xpruner} effectively reduces model size by removing a massive number of parameters without compromising performance.
Studies on applying LoRA~\citep{zhang2024loraprune,He2024VLMprune} or quantization~\citep{Guo2024jointprunequant} on top of pruned models have been conducted to further enhance the performance and efficiency of pruning strategies. 
\jwl{Although effective, most existing techniques assume unimodal models, limiting their effectiveness in multimodal settings.} For example, in \Cref{fig:multimodal_tok_one_col}, our empirical examination shows that conventional unimodal pruning methods, such as Wanda~\citep{Sung2024wanda} and similar pruning approaches~\citep{Zhang2024wanda2,He2024VLMprune,Sung2024ecoflap,Yin2024owl}, fail to generalize to multimodal settings where there can be substantial variances in input token activation and output token distributions across modalities~\citep{liang2024foundations}



Drawing inspiration from these observations, we introduce \textbf{T}oken-\textbf{A}daptive \textbf{M}ultimodal \textbf{P}runing (TAMP), a novel MLLM pruning framework that leverages inherent multimodal token attributes. \ours comprises two key components: first, we employ a layer-wise sparsity ratio strategy that dynamically adjusts the sparsity ratio per layer, \jwl{guided by the varying output token distributions.}
Specifically, \jwl{we assign lower sparsity ratios to layers exhibiting greater output token variations}, ensuring that these layers retain sufficient parameters to encode \jwl{rich} multimodal representations.
\jwl{Second,} instead of using all input tokens to compute input activations, we utilize attention scores to identify key multimodal input tokens that account for 
\jwl{each layer's unique multimodal processing demands.}

We validate our approach in various pruning scenarios using two distinct MLLMs, LLaVA-NeXT~\citep{Li2024llavanext} and VideoLLaMA2~\citep{Cheng2024videollama2}, evaluated on diverse multimodal benchmarks. Our layer-wise sparsity ratio strategy, based on varying output distributions, alone outperforms recent layer-wise sparsity approaches like ECoFLAP~\citep{Sung2024ecoflap} and OWL~\citep{Yin2024owl}, with 4.0\% higher performance at 50\% sparsity. Moreover, our approach of selecting multimodal tokens for input activations achieves up to 4.1\% performance gains over the state-of-the-art LLM pruning method Wanda~\citep{Sung2024wanda} at 50\% sparsity. \jwl{Combining both strategies further enhances performance, consistently surpassing strong pruning baselines. Our approach shows robustness at high sparsity, where our approach outperforms the second-best baseline with 8.2\% higher performance at 70\% sparsity.} Notably, our approach exclusively uses multimodal token attributes, avoiding the need for resource-intensive gradient or Hessian computations~\citep{Frantar2023sparsegpt,Sung2024ecoflap}, \jwl{supporting its efficiency by leveraging multimodal attributes for effective MLLM pruning.} 

\jwl{In summary, our contributions are as follows:}
\begin{itemize}[itemsep=0.0mm, parsep=1pt]
\item \jwl{We conduct comprehensive analyses and ablation studies to identify the importance of multimodal tokens in MLLM pruning. These include extensive analyses of multimodal token distributions across layers and in-depth investigations into their impact on pruning.}
\item \jwl{We introduce \ours, an effective MLLM pruning pipeline that leverages multimodal token attributes to measure layer importance for layer-wise sparsity and computes adaptive input activations for capturing multimodal processing demands at each layer.} 
\item \jwl{We validate our method on MLLMs that reflect their latest trends, demonstrating its effectiveness in preserving diverse multimodal abilities. Ours consistently outperforms pruning baselines, even at extreme pruning ratios.}

\end{itemize}

\section{Related Work}
\label{sec:related_work}
\begin{figure*}[t]
    \centering
    \includegraphics[width=\linewidth]{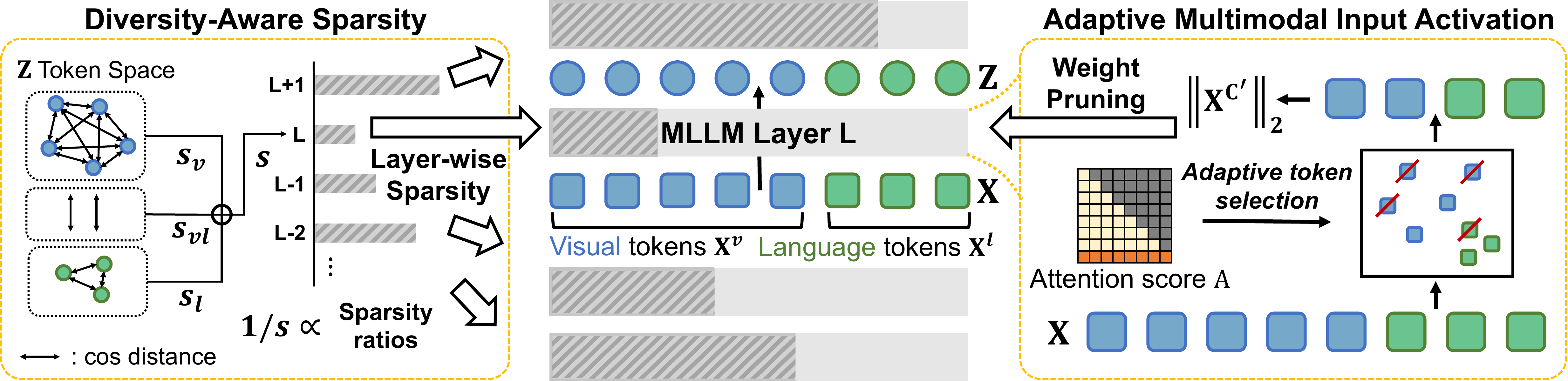}
    \par
    \caption{\small Overview of \ours. Our method utilizes multimodal token attributes to guide MLLM pruning. (\textbf{Left}): To effectively preserve each MLLM layer's differing capability to encode rich multimodal output tokens after pruning, we apply layer-wise sparsity, assigning sparsity inversely to the layer's importance, which is computed as the average of intra-modality ($\mathbf{s}_{v}, \mathbf{s}_{l}$) and inter-modality ($\mathbf{s}_{vl}$) diversities (\Cref{sec:sub:diversity-aware-sparsity}). (\textbf{Right}): To capture unique multimodal processing demands across different layers, we leverage attention scores to adaptively select multimodal input tokens for input activation calculations (\Cref{sec:sub:adaptive-selection}).
    }
    \label{fig:overview}
\end{figure*}

\paragraph{Multimodal Large Language Models} 
Recent advancements in Large Language Models (LLMs), such as LLaMA~\citep{llama,llama2}, Qwen~\citep{qwen2}, and \jwl{DeepSeek~\citep{deepseek}} have achieved remarkable progress in various natural language processing tasks by scaling to billions of parameters.
\jwl{Building on this success, Multimodal Large Language Models (MLLMs) have emerged as a new standard of multimodal models}, integrating multiple modalities, including text, image, audio and video, into a unified framework to address complex multimodal challenges in the real world \citep{Li2024llavaonevision,Zhan2024anygpt,Wu2024nextgpt}. 

LLaVA-NeXT~\citep{Li2024llavanext} integrates a visual encoder into LLMs and facilitates the understanding of high-resolution images, improving tasks such as visual question answering~\citep{Masry2022chartqa,Kembhavi2016ai2d} and visual reasoning~\citep{Yue2024mmmu,Liu2024mmbench}. Expanding beyond images, MLLMs such as VideoLLaMA2~\citep{Cheng2024videollama2} and LLaVA-OneVision~\citep{Zhan2024anygpt} have broadened their potential applications by incorporating other modalities such as audio, video, and interleaved images. However, as MLLMs continue to grow in size, their deployment in resource-constrained environments becomes increasingly challenging.

\paragraph{Model compression}
To tackle the challenges posed by increasing model scale, model compression techniques have emerged as a critical research area, aiming to optimize model size while maintaining performance~\cite{yao2022zeroquant,Wang2024QVLM,Frantar2023gptq,Lin2024AWQ}. Among these techniques, model pruning has gained prominence by removing redundant parameters or structures that minimally contribute to overall performance~\citep{Sung2024wanda, structured_sparsity, Ma2023llmpruner}. Approaches like Wanda \citep{Sung2024wanda} utilize weight magnitudes and input activations to compress LLMs, while SparseGPT \citep{Frantar2023sparsegpt} addresses the challenge of LLM pruning from the perspective of layer-wise output reconstruction problem.

However, all the aforementioned works are primarily designed for unimodal models, limiting their applicability to MLLMs. While ECoFLaP~\citep{Sung2024ecoflap} and VLMPrune~\citep{He2024VLMprune} extend pruning strategies to Vision-Language Models (VLMs) by applying layer-wise sparsity ratios tailored to vision-language characteristics, they treat multimodal tokens as if they originate from a single modality, overlooking their unique properties. In contrast, our work examines the impact of multimodal tokens on MLLM weight pruning and explicitly leverages multimodal properties for optimal pruning. Unlike prior approaches, we conduct comprehensive experiments on recent MLLMs, including those with more than two modalities, aligning with the latest advancements in MLLM design.

\section{Method}
In this section, we first present empirical studies that reveal key 
\jwl{properties of multimodal tokens and their implications for pruning.}
Based on these insights, we introduce the core components of Token Adaptive Multimodal Pruning (TAMP). The overall framework is illustrated in~\Cref{fig:overview}.
 
\begin{figure*}[t]
    \centering
    \includegraphics[width=\linewidth]{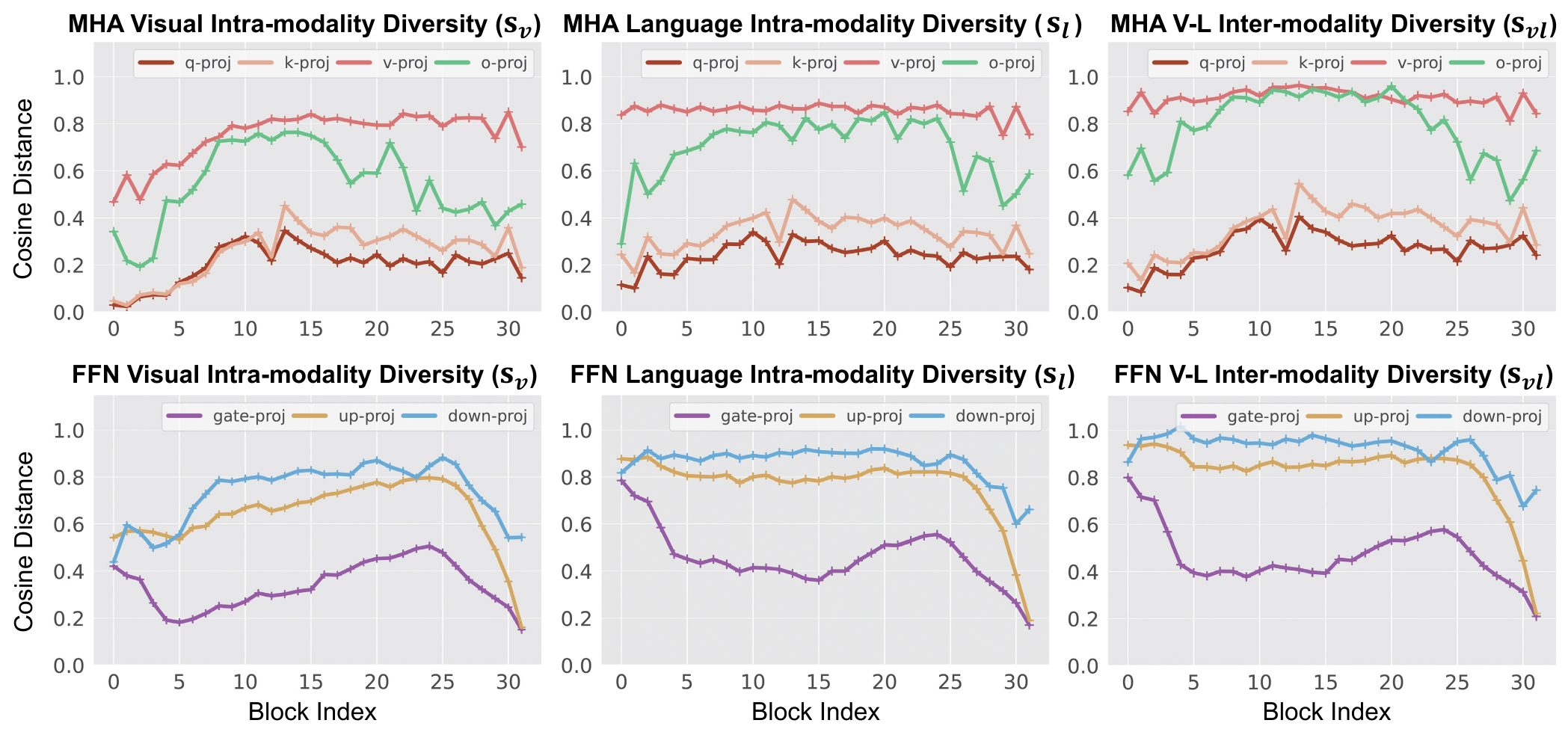}
    \par
    \caption{\small Intra-modality diversities ($\mathbf{s}_v, \mathbf{s}_l$) measure the average cosine distances among output tokens within the same modality, and inter-modality diversity ($\mathbf{s}_{vl}$) measures distances between output tokens from different modalities. We compute these diversities for each projection type in multi-head attention (Top) and feed-forward network (Bottom) across LLaVA-NeXT blocks using 128 randomly sampled inputs from the calibration set (ShareGPT4V). Notably, diversity trends differ by (1) modalities, (2) projection types, and (3) blocks, demonstrating varying capacities that should be preserved to effectively encode multimodal information across layers.} 
    \label{fig:similarity_trend}
\end{figure*}
\subsection{Preliminaries}\label{sec:sub:prelim}
A predominant MLLM typically consists of modality-specific encoders connected to an LLM through intermediate networks, with multimodal information from these encoders provided to the LLM as input tokens. While the following descriptions focus on an MLLM that uses an image encoder for visual information for simplicity, our approach is extensible to MLLMs that process other modalities, such as audio, video, or both.

Each block of the LLM processes two types of input tokens: visual $\mathbf{X}^{v}\!\in\!\mathbb{R}^{N_v\times C_{in}}$ and language $\mathbf{X}^{l}\!\in\!\mathbb{R}^{N_l\times C_{in}}$ input tokens, where $N_v$ and $N_l$ denote their respective token counts, and $C_{in}$ is the input dimension size. The block contains a multi-head attention \jwl{(MHA)} module, which computes an attention score $\mathbf{A}\!\in\!\mathbb{R}^{(N_v+N_L)\times (N_v+N_L)}$ that measures interplay between tokens, and a feed-forward network (FFN) module, which refines the output from the MHA module. 
Within these modules, varying types of linear projection layers $\mathbf{W}\!\in\!\mathbb{R}^{C_{out}\times C_{in}}$ transform input tokens into output tokens, where $C_{out}$ is the output dimension size: 
\jwl{
\begin{equation}
    \mathbf{Z}=\begin{bmatrix} \mathbf{Z}^{v} \\ \mathbf{Z}^{l} \end{bmatrix}=\begin{bmatrix} \mathbf{X}^{v} \\ \mathbf{X}^{l} \end{bmatrix}\mathbf{W}^{\top}=\mathbf{X}\mathbf{W}^{\top},
    \label{eq:output_token}
\end{equation}
}
where $\mathbf{Z}^{v}$ and $\mathbf{Z}^{l}$ represent the visual and language output tokens, respectively. To determine which parameters to prune, many predominant methods\jwl{~\citep{Sung2024wanda,Sung2024ecoflap,Yin2024owl}}
\jwl{define layer's parameter importance based on input activation and weight magnitude, computed as: $\mathbf{I}=||\mathbf{X}||_{2}\cdot|\mathbf{W}|$, where $|\cdot|$ is element-wise absolute value operator and $||\mathbf{X}||_{2}\!\in\!\mathbb{R}^{C_{in}}$ is the input activation computed for each channel as $\ell_{2}$-norm of that channel's activation across all input tokens. Parameters with the lowest importance are considered redundant and thus pruned.}

\subsection{Diversity-Aware Sparsity}\label{sec:sub:diversity-aware-sparsity}
\jwl{We first conduct a systemic study of the distributional properties of multimodal output tokens} by computing intra- and inter-modality diversities. \jwl{Intra-modality diversities measure the distances among output tokens within the same modality, while inter-modality diversity quantifies those between output tokens from different modalities:}
\begin{align}
    \hspace{-2mm}
    \begin{split}
    \mathbf{s}_{v}&=\mathbb{E}_{i,j\sim\mathcal{C}_v}\left[\mathbf{d}_{ij}\right],\;\mathbf{s}_{l}=\mathbb{E}_{i,j\sim\mathcal{C}_l}\left[\mathbf{d}_{ij}\right],\\
    \mathbf{s}_{vl}&=\mathbb{E}_{i\sim\mathcal{C}_v,\;j\sim\mathcal{C}_l}\left[\mathbf{d}_{ij}\right],\;\jwl{\mathbf{d}_{ij}=1-\langle\mathbf{Z}_i,\mathbf{Z}_j\rangle},
    \label{eq:inter_intra_dist}
    \end{split}
    \hspace{-2mm}
\end{align}
where $\mathcal{C}_v$ and $\mathcal{C}_l$ denote visual and language token indices, respectively, and $\mathbf{d}_{ij}$ is \jwl{the cosine distance between output tokens. $\mathbf{s}_{v}$ and $\mathbf{s}_l$ are intra-modality diversities of visual and language modalities, respectively, while $\mathbf{s}_{vl}$ is inter-modality diversity.}

\Cref{fig:similarity_trend} illustrates these diversities across projection layer types and MLLM blocks.
\jwl{We observe three key properties of output token distributions: (1) Comparing $\mathbf{s}_{v}$, $\mathbf{s}_{l}$ and $\mathbf{s}_{vl}$ reveals notable difference across modalities. This confirms the need to compute intra- and inter-modality diversities separately to capture unique patterns of each modality; (2) both intra- and inter-modality diversities vary significantly across layer types.}
\jwl{As shown in~\Cref{fig:similarity_trend} \highlight{Top},} value projection layers (v-proj) exhibit higher diversities than query (q-proj) and key (k-proj) projection layers, despite receiving identical input tokens within the same block, suggesting that value layers encode richer multimodal information; \jwl{(3) these diversities fluctuate significantly across blocks, indicating that the capability to encode multimodal information varies with model depth.}

\jwl{To address these observations in MLLM pruning, we propose a layer-wise sparsity strategy based on multimodal output token diversity.} Our core intuition is that layers with higher multimodal output token diversity should retain more parameters \jwl{during pruning} to maintain their capability to encode \jwl{richer} multimodal output tokens. We quantify the importance of each MLLM layer as the average of intra- and inter-modality diversities: $\mathbf{s}=(\mathbf{s}_v+\mathbf{s}_l+\mathbf{s}_{vl})/3$. Following ECoFLAP~\citep{Sung2024ecoflap}, \jwl{sparsity is set inversely proportional to the precomputed layer's importance}, ensuring that layers with higher diversities retain more parameters to preserve their representational capabilities. 

\subsection{Influence of Multimodal Input Tokens}\label{sec:sub:empirical}
We now shift our focus to the impact of input tokens in MLLM pruning. \jwl{Our primary assumption is that input tokens from different modalities contribute distinctively to multimodal information processing.} 
To investigate this, we first analyze attention distributions across blocks for each modality by computing the average attention scores per modality from the attention score matrix $\mathbf{A}$, as depicted in~\Cref{fig:attention_trend}. The result reveals a clear trend: different blocks put varying degrees of reliance on visual and language inputs. This variation in modality reliance across blocks implies that a static, uniform input activation calculation may be suboptimal. 

This phenomenon \jwl{motivates us to examine whether modality-specific input tokens contribute distinctively to pruning outcomes.
To explore this, we conduct preliminary experiments comparing two approaches for computing input activations: (1) the conventional approach of using both visual and language input tokens ($||\mathbf{X}||_{2}$, denoted as "V+L"), and (2) a variant that focuses solely on language tokens ($||\mathbf{X}^{l}||_{2}$, denoted as "L"). We apply both methods across the 32 blocks of LLaVA-NeXT~\citep{Li2024llavanext} to assess the influence of multimodal tokens at different network depths.
} 


\Cref{tab:token_selection_ablation} presents the pruning results for the conventional approach (V+L) alongside the variant (L). Notably, including both visual and language tokens across all blocks achieves better performance in a visually rich information understanding task (e.g., ChartQA). In contrast, omitting visual tokens reduces performance on ChartQA but improves results on a multimodal understanding task (e.g., MME). \jwl{These results confirm our intuition that different blocks engage with specific modalities to varying degrees for multimodal information processing, which influences pruning outcomes.}

\subsection{Adaptive Multimodal Input Activation}\label{sec:sub:adaptive-selection}
The above findings support the need for a pruning strategy that dynamically adapts to modality-specific contributions of individual blocks. To address this, we propose an adaptive method that selects multimodal input tokens for input activation calculations tailored to \jwl{address} each block's unique multimodal processing needs.

\jwl{A key step in our approach is identifying core input tokens by measuring their contributions.} In this work, we use the last row of the attention score matrix $\mathbf{A}$ as token contributions: \jwl{$\mathbf{a}\!=\!\mathbf{A}[:,\!-\!1]\!\in\!\mathbb{R}^{N_v+N_l}$}, which captures importance of multimodal tokens. This can guide the dynamic selection of input tokens based on the unique processing demand of each block. For example, in a layer emphasizing visual information, visual tokens have high contributions $\mathbf{a}$. \jwl{Thus, more visual tokens are prioritized during the selection for that layer, ensuring that its input activation retains crucial visual features.} 

For input token selection, we adopt the data selection algorithm in~\citet{Maharana2023d2prune} to select core input tokens while considering token diversity. This selection process prioritizes input tokens with high $\mathbf{a}$ values while ensuring that the output tokens they produce remain diverse in output token space. Specifically, we first update $\mathbf{a}$ by incorporating both the intrinsic and neighboring token contributions:
\begin{equation}
    \hspace{-2mm}
    \mathbf{a}_{i}\!\leftarrow\!\mathbf{a}_i\!+\!\sum_{j\in\mathcal{N}_i}\mathbf{e}_{ij}\!\cdot\!\mathbf{a}_{j},\;\;\mathbf{e}_{ij}\!=\!\exp\left(-\gamma*\mathbf{d}_{ij}\right),
    \label{eq:forward_pass}
    \hspace{-2mm}
\end{equation}
where $\mathcal{N}_i$ denotes \jwl{$i$th token}'s nearest neighbors. We consider three nearest neighbors and $\gamma=1$, following the default setting of the algorithm.

\begin{table}[t]
\centering
\begin{minipage}{\linewidth}
    \centering
    \includegraphics[width=0.9\linewidth]{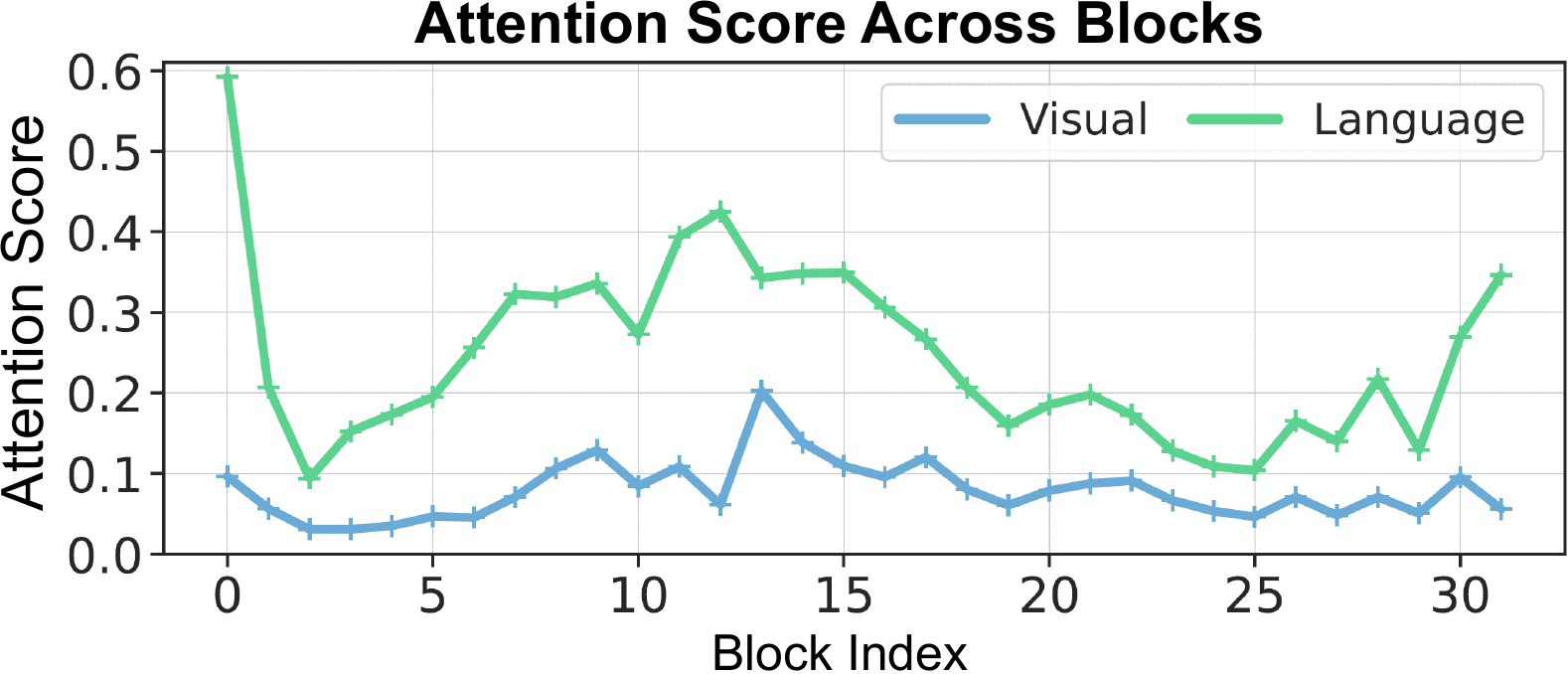}
    \par
    \captionof{figure}{\small Average attention score across LLaVA-NexT blocks. Varying attention scores indicate that unique multimodal processing demands exist for each block.}
    \label{fig:attention_trend}
\end{minipage}
\par
\vspace{0.1in}
\begin{minipage}{\linewidth}
    \captionsetup{type=table}
        \centering
        \resizebox{\linewidth}{!}{
            \renewcommand{\arraystretch}{1.2}
            \renewcommand{\tabcolsep}{1.5pt}
            \begin{tabular}{l c c c c}
                 \toprule
                 {\textbf{Method}} & {\textbf{MME-}} & {\textbf{MME-}} & {\textbf{ChartQA}} \\
                 & {\textbf{cognition}} & {\textbf{perception}} & \\
                 \midrule
                 Full Model &
                 {376.8} & {1588.3} & {69.2}\\
                  \cmidrule{0-4}
                 V+L (Block 1 - 32) &
                 {276.4} & {1360.6} & \textbf{63.2}\\
                 V+L (Block 1 - 2), L (Block 3 - 32) &
                 {\textbf{320.4}} & {\textbf{1476.6}} & {62.3}\\
                 L (Block 1 - 32) &
                 {311.1} & {1468.4} & {62.2}\\
                 \bottomrule
            \end{tabular}
        }
        \captionof{table}{\small Impact of token selection on 50\% pruning of LLaVA-NeXT across evaluation benchmarks. MME measures general multimodal understanding, while ChartQA focuses on visually rich information understanding (e.g., OCR, chart).}
        \label{tab:token_selection_ablation}
\end{minipage}
\end{table}
\jwl{Once updated, we iteratively select the token with the highest contribution. To encourage selection diversity}, the contributions of neighboring tokens of the selected token are penalized:
\begin{equation}
    \mathbf{a}_{j}\leftarrow \mathbf{a}_{j} - \mathbf{e}_{ij}\cdot\mathbf{a}_{i},\;\;\forall j\in\mathcal{N}_i,
    \label{eq:reverse_pass}
\end{equation}
where $\gamma=0.2$, following the original setting. \jwl{These iterative processes} prioritize core multimodal tokens while minimizing redundant selections.

We select tokens from the full token index set $\mathcal{C}$ until the selected index set $\mathcal{C}^{\prime}$ sufficiently represents the original distribution using maximum mean discrepancy (MMD) metric~\citep{Kim2016mmdcritic}:
\begin{align}
\text{MMD}\!=\!A(\mathcal{C},\mathcal{C})&\!+\!A(\mathcal{C}^{\prime},\mathcal{C}^{\prime})\!-\!2A(\mathcal{C},\mathcal{C}^{\prime})<0.1\!*\!\sqrt{\mathbf{s}},  \notag \\
    A(\mathcal{C},\mathcal{C}^{\prime})&=\frac{1}{|\mathcal{C}||\mathcal{C}^{\prime}|} \sum_{i \in \mathcal{C},\; j \in \mathcal{C}} \mathbf{e}_{ij}, \label{eq:mmd_select}
\end{align}
\jwl{where $A(\mathcal{C},\mathcal{C}^{\prime})$ measures the distributional similarity between two sets. The selection process continues until MMD falls below a threshold scaled by the function of ${\mathbf{s}}$, which accounts for variations in output token spaces, as shown in~\Cref{fig:similarity_trend}.} Input activation is then computed using the selected tokens: $||\mathbf{X}^{\mathcal{C}^{\prime}}||_{2}$. This approach adaptively captures multimodal processing demands across different layers, \jwl{which can facilitate pruning decisions by preserving parameters critical to those demands.}

\section{Experiments}\label{sec:experiments}

\subsection{Experimental Setups}\label{sec:experiments:datasets}

\paragraph{Multimodal Large Language Models} We conduct pruning on two popular MLLM architectures. LLaVA-NeXT~\citep{Li2024llavanext} 
with 8B parameters enhances visual perception by splitting high-resolution images into sub-images. VideoLLaMA2~\citep{Cheng2024videollama2} 
with 7B parameters improves spatiotemporal modeling and audio processing, making it well-suited for video and audio tasks. These models enable comprehensive evaluation of pruning strategies across diverse multimodal settings. \jwl{A recent study~\citep{He2024VLMprune} shows that pruning only the LLM component in MLLMs achieves a better balance between performance and efficiency since LLMs are typically much larger than these encoders. Therefore, our experiments focus on pruning the LLM component of MLLMs.} 

\paragraph{Evaluation Benchmark} To assess performance after pruning, we evaluate their zero-shot capability on various multimodal benchmarks. We follow the evaluation protocols outlined in LLaVA-NeXT and VideoLLaMA2 to ensure consistent benchmark selection. For LLaVA-NeXT, we evaluate its zero-shot performance on multiple vision-language tasks: 1) multimodal understanding: MME~\citep{Fu2023MME} and MMMU~\citep{Yue2024mmmu}; 2) visual mathematic reasoning: MathVista~\citep{Lu2024mathvista}; 3) structural reasoning: ChartQA~\citep{Masry2022chartqa} and AI2D~\citep{Kembhavi2016ai2d}; 4) multimodal perception: MMBench~\citep{Liu2024mmbench}. For VideoLLaMA2, we assess its performance across diverse multimodal settings: 1) audio: Clotho-AQA~\citep{Lipping2022clothoaqa} for open-ended QA, TUT2017~\citep{Mesaros2016TUT} and VocalSound~\citep{Gong2022vocalsound} for multiple-choice QA, and Muchomusic~\citep{Weck2024muchomusic} for music understanding; 2) video: VideoMME and NeXTQA-MC for diverse video domains and durations, EgoSchema for long video understanding, and MVBench for spatio-temporal understanding; 3) audiovisual comprehension: MUSIC-QA~\citep{Li2022musicqa} for open-ended musical scene understanding. Further details on the evaluation pipeline are provided in~\Cref{appendix:Implementation Details}.

\jwl{We report performances at sparsity ratios where pruned models maintain reasonably high performance that enables meaningful comparisons with baselines.} To ensure fair comparisons across benchmarks with different scales, we compute the average relative performance, denoted as \relp, which measures model generalization. We compute relative performance as: (pruned model performance / full-model performance) $\times$ 100\%. 

\paragraph{Baselines}
We compare our method with several widely used pruning approaches.  
Magnitude~\citep{zhu2017prune}, a standard baseline, removes weights with the smallest absolute values.  
SparseGPT~\citep{Frantar2023sparsegpt} is a layer-wise pruning method that leverages Hessian-based approximations to preserve critical weights.  
Wanda~\citep{Sung2024wanda} computes a layer-wise importance score as the product of weight magnitudes and input activations.  
OWL~\citep{Yin2024owl} proposes an outlier-weighted sparsity strategy, adjusting pruning ratios per layer based on outlier prevalence.
ECoFLaP~\citep{Sung2024ecoflap} uses zeroth-order gradient calculations to estimate the global importance score of VLM layers and determines layer sparsity ratios based on this score.

\subsection{Results and Discussion}\label{sec:experiments:result}
\vspace{-0.05in}
\begin{table*}[t]
    \tiny
    \centering
    \resizebox{\textwidth}{!}{
        \renewcommand{\arraystretch}{1.2}
        \renewcommand{\tabcolsep}{7.5pt}
        \begin{tabular}{l c c c c c c c a}
             \toprule
             {\textbf{Method}} & {\textbf{MME-}} & {\textbf{MME-}} & {\textbf{ChartQA}} & {\textbf{AI2D}} & {\textbf{MMMU}} & {\textbf{Mathvista}} & {\textbf{MMBench}} & {\textbf{\relp (\%)}}\\
              & {\textbf{cognition}} & {\textbf{perception}} & & & & & & \\
             \midrule
             Full Model &
             {\scriptsize 376.8} & {\scriptsize 1588.3} & {\scriptsize 69.2} & {\scriptsize 71.7} & {\scriptsize 40.1} & {\scriptsize 36.2} & {\scriptsize 72.2} & {\scriptsize 100} \\
             \cmidrule{0-8}
             Magnitude &
             {\scriptsize 0} & {\scriptsize 0} & {\scriptsize 0} & {\scriptsize 0} & {\scriptsize 24.0} & {\scriptsize 26.6} & {\scriptsize 0} & {\scriptsize 19.0} \\
              SparseGPT &
             {\scriptsize \underline{328.6}} & {\scriptsize \underline{1448.9}} & {\scriptsize \textbf{65.5}} & {\scriptsize 64.5} & {\scriptsize 33.6} & {\scriptsize \underline{31.3}} & {\scriptsize 64.7} & {\scriptsize \underline{89.0}} \\
                Wanda &
             {\scriptsize 276.4} & {\scriptsize 1360.6} & {\scriptsize 63.2} & {\scriptsize 64.3} & {\scriptsize \textbf{36.2}} & {\scriptsize 30.2} & {\scriptsize 63.9} & {\scriptsize 86.0} \\
              ECoFLaP &
             {\scriptsize 254.6} & {\scriptsize 1429.5} & {\scriptsize \textbf{65.5}} & {\scriptsize \textbf{66.1}} & {\scriptsize 35.1} & {\scriptsize 30.7} & {\scriptsize \underline{66.2}} & {\scriptsize 86.9} \\
              OWL &
             {\scriptsize 274.3} & {\scriptsize 1366.0} & {\scriptsize 63.2} & {\scriptsize 64.0} & {\scriptsize 35.3} & {\scriptsize 30.8} & {\scriptsize 64.1} & {\scriptsize 85.9} \\
             \cellcolor{gg}\ours (Ours) &
              \cellcolor{gg}{\scriptsize \textbf{341.0}} & \cellcolor{gg}{\scriptsize \textbf{1470.2}} & \cellcolor{gg}{\scriptsize 64.7} & \cellcolor{gg}{\scriptsize \underline{65.0}} & \cellcolor{gg}{\scriptsize \underline{35.7}} & \cellcolor{gg}{\scriptsize \textbf{31.9}} & \cellcolor{gg}{\scriptsize \textbf{66.3}} & {\scriptsize \textbf{90.9}} \\
             \bottomrule
        \end{tabular}
    }
    \caption{\small Comparison of pruning techniques on the LLaVA-NeXT model with 50\% sparsity ratio and estimate performance on various multimodal evaluation benchmarks. The best and the second best results are in \textbf{bold} and \underline{underlined}, respectively.}
    \label{tab:llava_next_eval}
\end{table*}

\begin{table*}[t]
    \tiny
    \centering
    \resizebox{\textwidth}{!}{
        \renewcommand{\arraystretch}{1.2}
        \renewcommand{\tabcolsep}{6.5pt}
        \begin{tabular}{l|c c c c|c c c c|c a}
             \toprule
              & \multicolumn{4}{c|}{\textbf{Audio}} & \multicolumn{4}{c|}{\textbf{Video}} & {\textbf{Audiovisual}} & \\
             {\textbf{Method}} & {\textbf{Clotho}} & {\textbf{TUT}} & {\textbf{Vocal}} & {\textbf{Mucho}} & {\textbf{Video}} & {\textbf{Ego}} & {\textbf{NextQA}} & {\textbf{MV}} & {\textbf{MUSIC}} & {\textbf{\relp (\%)}} \\
              & {\textbf{-AQA}} & {\textbf{2017}} & {\textbf{Sound}} & {\textbf{music}} & {\textbf{MME}} & {\textbf{Schema}} & {\textbf{-MC}} & {\textbf{-Bench}} & {\textbf{-QA}} & \\
             \midrule
             Full Model &
             {\scriptsize 85.6} & {\scriptsize 71.2} & {\scriptsize 92.4} & {\scriptsize 58.9} & {\scriptsize 48.7} & {\scriptsize 49.3} & {\scriptsize 73.3} &{\scriptsize 58.4} & {\scriptsize 79.4} & {\scriptsize 100} \\
             \cmidrule{0-10}
             Magnitude &
             {\scriptsize 0} & {\scriptsize 0} & {\scriptsize 0} & {\scriptsize 25.8} & {\scriptsize 0} & {\scriptsize 20.8} & {\scriptsize 20.3} &{\scriptsize 0} & {\scriptsize 0} & {\scriptsize 12.6} \\
              SparseGPT &
             {\scriptsize \underline{83.9}} & {\scriptsize 64.1} & {\scriptsize 91.9} & {\scriptsize 48.8} & {\scriptsize 35.7} & {\scriptsize 42.6} & {\scriptsize 61.8} &{\scriptsize 54.2} & {\scriptsize 70.6} & {\scriptsize 88.5} \\
                Wanda &
             {\scriptsize 83.1} & {\scriptsize 65.6} & {\scriptsize \textbf{92.1}} & {\scriptsize 51.4} & {\scriptsize 39.4} & {\scriptsize 44.4} & {\scriptsize 65.0} &{\scriptsize 53.2} & {\scriptsize \textbf{73.3}} & {\scriptsize 91.0} \\
              ECoFLaP &
             {\scriptsize 83.7} & {\scriptsize 67.2} & {\scriptsize \textbf{92.1}} & {\scriptsize \underline{54.4}} & {\scriptsize \underline{41.3}} & {\scriptsize \textbf{46.8}} & {\scriptsize \underline{69.8}} &{\scriptsize \underline{54.2}} & {\scriptsize \underline{73.0}} & {\scriptsize \underline{93.8}} \\
              OWL &
             {\scriptsize 83.2} & {\scriptsize \textbf{70.5}} & {\scriptsize 91.3} & {\scriptsize 47.6} & {\scriptsize 37.7} & {\scriptsize 43.6} & {\scriptsize 63.1} &{\scriptsize 52.4} & {\scriptsize 68.9} & {\scriptsize 89.4} \\
             \cellcolor{gg}\ours (Ours) &
             \cellcolor{gg}{\scriptsize \textbf{84.2}} & \cellcolor{gg}{\scriptsize \underline{69.9}} & \cellcolor{gg}{\scriptsize \textbf{92.1}} & \cellcolor{gg}{\scriptsize \textbf{55.9}} & \cellcolor{gg}{\scriptsize \textbf{42.5}} & \cellcolor{gg}{\scriptsize \underline{46.7}} & \cellcolor{gg}{\scriptsize \textbf{70.9}} & \cellcolor{gg}{\scriptsize \textbf{54.8}} & \cellcolor{gg}{\scriptsize 72.6} & {\scriptsize \textbf{95.0}} \\
             \bottomrule
        \end{tabular}
    }
    \caption{\small Comparison of pruning techniques on the VideoLLaMA2 model with 60\% sparsity ratio and estimate performance on various multimodal evaluation benchmarks. The best and the second best results are in \textbf{bold} and \underline{underlined}, respectively.}
    \vspace{-0.1in}
    \label{tab:videollama2_eval}
\end{table*}

\paragraph{\ours outperforms baselines on LLaVA-NeXT.}
\Cref{tab:llava_next_eval} reports performance of LLaVA-NeXT at a 50\% sparsity ratio.
Across 6 of 7 benchmarks, including MME, AI2D, MMMU, Mathvista, and MMBench, \ours ranks either first or second. On average, \ours surpasses the strongest baseline by 1.9 percent points (pp) in relative performance, demonstrating its strength in preserving key parameters essential for versatile visual comprehension. 

\begin{figure}[t]
\centering
    \centering
    \includegraphics[width=0.95\linewidth]{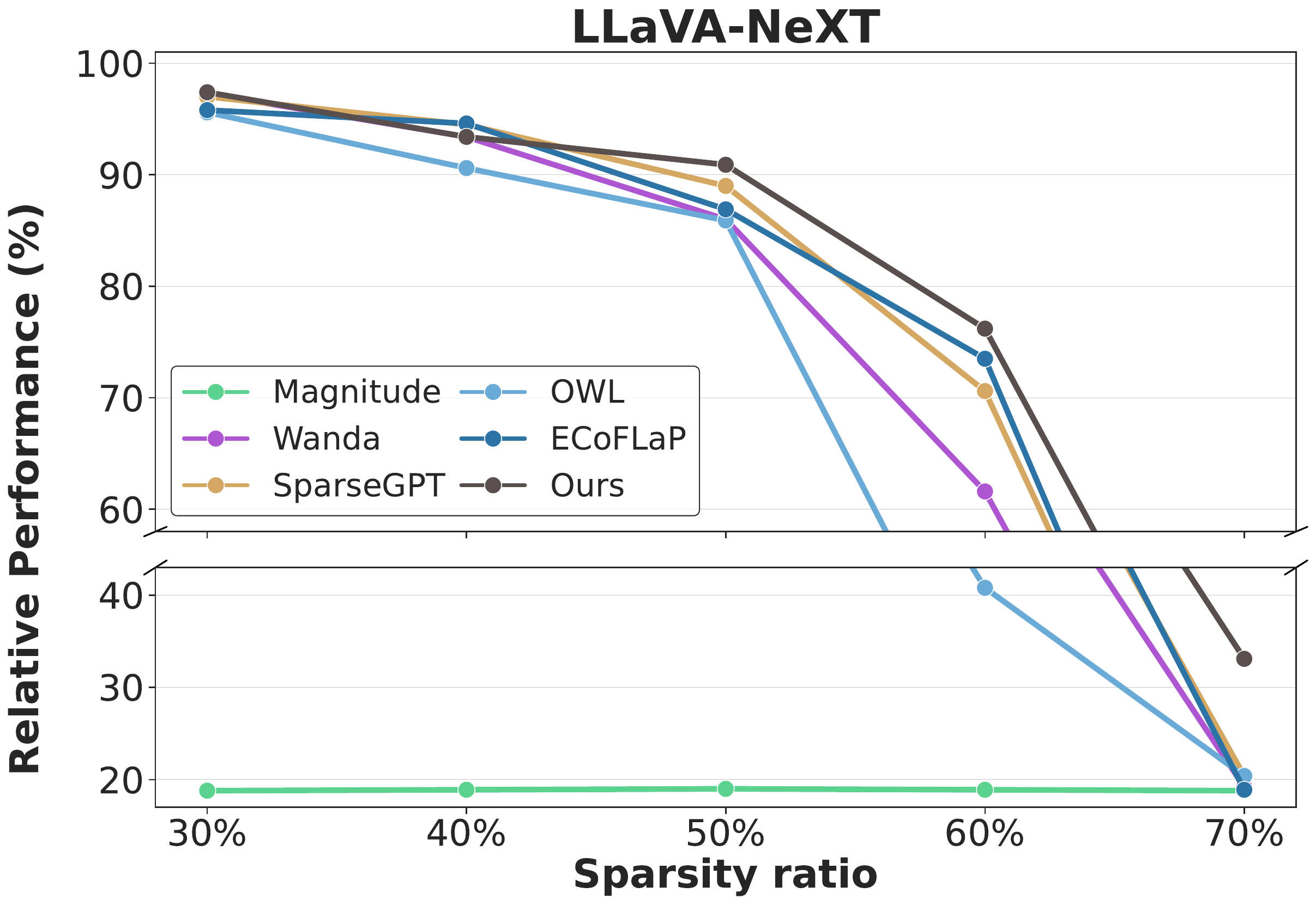}
    \par
    \caption{\small Average relative performances of all pruning techniques at different sparsity ratios for the LLaVA-NeXT.}
    \label{fig:llava_next_sparsity_ratio}
\vspace{-0.15in}
\end{figure}

Furthermore, \Cref{fig:llava_next_sparsity_ratio} presents the performance of LLaVA-NeXT across a range of sparsity levels. \ours exhibits the best performance-sparsity trade-off. In contrast, pruning baselines experience steep accuracy declines beyond 50\% sparsity, whereas our adaptive approach shows superior retention of model ability in high sparsity regimes (e.g., 60\% and 70\%), highlighting its robustness.
\vspace{-0.05in}

\paragraph{\ours effectively preserves diverse multimodal understanding.} To further examine our approach, we evaluate VideoLLaMA2 at a 60\% sparsity ratio, with results presented in~\Cref{tab:videollama2_eval}. \ours ranks the top position in nearly all audio and video tasks and a close second in the audiovisual benchmark, outperforming the second-best baseline by 1.2 pp in average relative performance. These results demonstrate that our approach effectively captures modality-specific contributions, validating its universality across multiple modalities and tasks. Additional experiments on LLaVA-OneVision~\citep{Li2024llavaonevision}, which handles interleaved image and video modalities, are provided in~\Cref{appendix:LLaVA-Onevision_exp}.

\begin{figure}[t]
    \centering
     \includegraphics[width=0.95\linewidth]{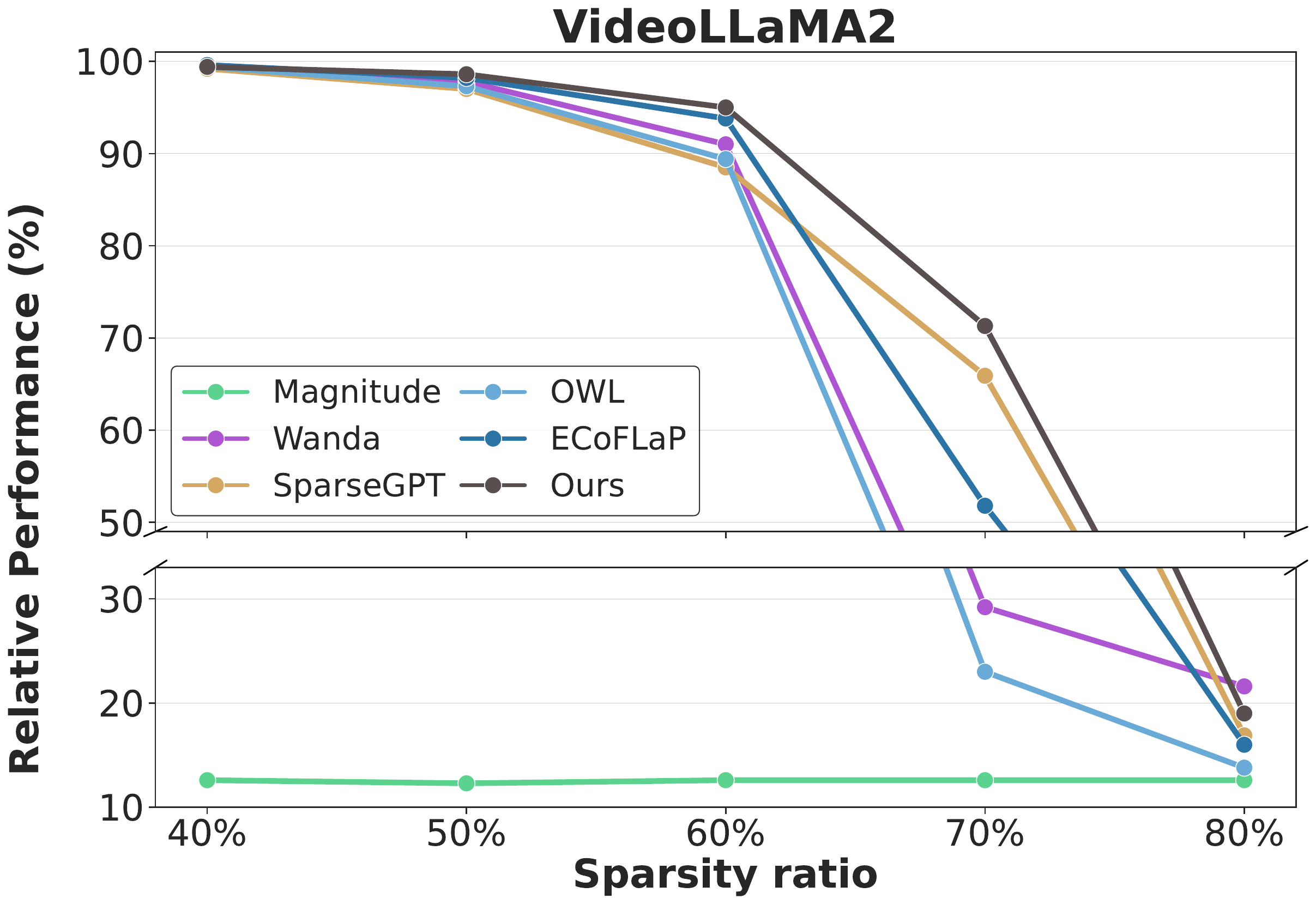}
     \par
    \caption{\small Average relative performances of all pruning techniques at different sparsity ratios for the VideoLLaMA2.}
    \label{fig:videollama2_sparsity_ratio}
    \vspace{-0.15in}
\end{figure}
As shown in~\Cref{fig:videollama2_sparsity_ratio}, \ours consistently maintains strong performance across different sparsity levels in VideoLLaMA2. This further shows \ours's robustness in maintaining diverse multimodal comprehension even under aggressive sparsity constraints. Moreover, in both~\Cref{fig:llava_next_sparsity_ratio} and~\Cref{fig:videollama2_sparsity_ratio}, OWL suffers from severe performance drops at high sparsity ratios, unlike ECoFLaP and \ours. OWL assigns layer-wise sparsity ratios proportional to the prevalence of outlier values within input activations computed across all input tokens. We hypothesize that multimodal encoder's tokens in MLLMs follow different outlier distributions than unimodal language tokens in LLMs, where large activation values in tokens are typically featured in LLMs~\citep{Sun2024massive_act,Yin2024owl}. This discrepancy likely contributes to OWL's underperformance, highlighting the importance of pruning strategies tailored for MLLMs to account for their unique multimodal attributes.
\begin{table*}[t]
\centering
\vspace{-0.075in}
    \begin{minipage}{0.37\textwidth}
        \centering
        \small{\textbf{(a) Key Components}}\\
        \vspace{0.025in}
        \resizebox{\linewidth}{!}{%
\renewcommand{\arraystretch}{1.25}
\renewcommand{\tabcolsep}{2.5pt}
\begin{tabular}{lcc c c}
    \toprule
    \textbf{Method} & \textbf{DAS} & \textbf{AMIA} & \textbf{LLaVA} & \textbf{Video} \\
    &  &  & \textbf{-NeXT} & \textbf{LLaMA2} \\
    \midrule
    Wanda & $-$ & $-$ & 86.0 & 91.0 \\ 
    ECoFLaP & $-$ & $-$ & 86.9 & 93.8 \\
    OWL & $-$ & $-$ & 85.9 & 89.4 \\
    \midrule
    \multirow{3}{*}{{\ours (Ours)}} & $\checkmark$ & $-$ & \underline{90.4} & \underline{94.0} \\
    & $-$ & $\checkmark$ & 89.5 & 92.3 \\
    & $\checkmark$ & $\checkmark$ & \textbf{90.9} & \textbf{95.0} \\
    \bottomrule
\end{tabular}}
    \end{minipage}
    \begin{minipage}{0.32\textwidth}
        \centering
        \small{\textbf{(b) Layer-wise Sparsity}}\\
        \vspace{0.025in}
        \resizebox{\textwidth}{!}{
    \renewcommand{\arraystretch}{1.325}
    \renewcommand{\tabcolsep}{3.0pt}
    \begin{tabular}{l c c}
         \toprule
         {\textbf{Method}} & {\textbf{LLaVA}} & {\textbf{Video}}\\
          & {\textbf{-NeXT}} & {\textbf{LLaMA2}}\\
         \midrule
         Wanda & 86.0 & 91.0 \\
         + All-token DAS & 89.3 & 94.1 \\
         + Block-wise DAS & 88.0 & 94.2 \\
         + DAS (Ours) & 90.4 & 94.0 \\
         SparseGPT & 89.0 & 88.5 \\
         + DAS (Ours) & 89.1 & 94.0 \\
         \bottomrule
    \end{tabular}
}
    \end{minipage}
    \begin{minipage}{0.29\textwidth}
        \centering
        \small{\textbf{(c) Input Activation}}\\
        \vspace{0.025in}
        \resizebox{\textwidth}{!}{
    \renewcommand{\arraystretch}{1.7}
    \renewcommand{\tabcolsep}{3.0pt}
    \begin{tabular}{l c c}
         \toprule
         {\textbf{Method}} & {\textbf{LLaVA}} & {\textbf{Video}}\\
         & {\textbf{-NeXT}} & {\textbf{LLaMA2}}\\
         \midrule
         Wanda & 86.0 & 91.0 \\
         Random & 85.3 & 91.1 \\
         Attention & 89.2 & 93.5 \\
         AMIA (Ours) & 89.5 & 92.3 \\
         \bottomrule
    \end{tabular}
}
    \end{minipage}
\caption{\small Ablation studies of \ours. DAS: Diversity-Aware Sparsity in~\Cref{sec:sub:diversity-aware-sparsity}, AMIA: Adaptive Multimodal Input Activation in~\Cref{sec:sub:adaptive-selection}. (a) Contributions of proposed components. (b) Ablation on layer-wise sparsity strategies. (c) The performance of different multimodal input token selections for input activations calculation. For all experiments, we prune LLaVA-NeXT at 50\% and VideoLLaMA2 at 60\% sparsity ratios, and report the relative average performance.}
\label{tab:components_study}
\end{table*}

\subsection{Further Analysis and Ablation \label{sec:subsec:analysis}}
\paragraph{Core components in \ours contribute to improving performance.}
To validate our strategies, we compare the two core components of \ours, Diversity-Aware Sparsity (DAS) and Adaptive Multimodal Input Activation (AMIA), against ECoFLaP, OWL, and Wanda. Like DAS, ECoFLaP and OWL assign varying sparsity ratios to layers. AMIA selects core multimodal input tokens for input activations, while Wanda uses all input tokens. All the above methods build upon Wanda. 

As shown in~\Cref{tab:components_study} \highlight{(a)}, both DAS and AMIA bring substantial performance gain. DAS alone surpasses ECoFLaP and OWL, improving Wanda by 4.4 pp in LLaAV-NeXT and 3.0 pp in VideoLLaMA2. \jwl{This supports the importance of multimodal token diversity in identifying critical layers for encoding rich multimodal representation.} Notably, DAS outperforms ECoFLaP, which relies on gradient computations, \jwl{despite calculating simple cosine distances among tokens.} This demonstrates that DAS efficiently captures the complexities of multimodal data, further validating its efficacy.

AMIA improves Wanda by 3.5 pp in LLaVA-NeXT and 1.3 pp in VideoLLaMA2. This improvement stems from AMIA's adaptive selection of core multimodal tokens for input activations, \jwl{aligning pruning with each layer's processing needs.} Integrating DAS and AMIA, \ours achieves superior performance, underscoring the advantage of jointly optimizing layer-wise sparsity and pruning decisions in MLLMs through multimodal attributes. 

\paragraph{Ablation on layer-wise sparsity.}
We further test variants of DAS. All-token DAS averages the cosine distances of all output tokens to determine layer importance: $\mathbf{s}\!=\!\mathbb{E}_{i,j\sim\mathcal{C}}\left[\mathbf{d}_{ij}\right]$. Block-wise DAS averages the layer importance in DAS within each block and applies uniform sparsity to all layers in that block. The results are summarized in~\Cref{tab:components_study} \highlight{(b)}. DAS shows robust performance across MLLMs compared to All-token DAS, validating the use of intra- and inter-modality diversities to reflect unique token distributions across modalities. DAS also improves SparseGPT's performance, further demonstrating its adaptability across pruning methods. Block-wise DAS outperforms ECoFLaP and OWL, demonstrating that our approach can also represent block-level importance. To provide deeper insights, we further analyze the sparsity ratios of baselines and our approach in~\Cref{appendix:sparsity_analysis}. 

\paragraph{Adaptive Input Activation.} To further validate our intuition that MLLM pruning needs to adapt to modality-specific contributions within each block, we conduct ablation studies on different token selection strategies for input activations. Our approach, AMIA, selects core tokens based on token contribution score $\mathbf{a}$ and output token distances. We examine other selection strategies: (1) Random, which randomly selects 100 tokens, and (2) Attention, which selects tokens with above-average contribution scores. As shown in~\Cref{tab:components_study} \highlight{(c)}, attention-based selection methods, Attention and AMIA, outperform random selection, supporting our core intuition. However, the best strategy depends on the target MLLM. This may be due to the complexity of multimodal feature spaces, suggesting that further refinements in selection methods could enhance pruning robustness. We analyze AMIA's token selection with visualizations in~\Cref{appendix:token_selection}.

\begin{table*}[t]
    \tiny
    \centering
    \resizebox{\textwidth}{!}{
        \renewcommand{\arraystretch}{1.2}
        \renewcommand{\tabcolsep}{6.5pt}
        \begin{tabular}{l|c c c c|c c c c|c a}
             \toprule
              & \multicolumn{4}{c|}{\textbf{Audio}} & \multicolumn{4}{c|}{\textbf{Video}} & {\textbf{Audiovisual}} & \\
             {\textbf{Method}} & {\textbf{Clotho}} & {\textbf{TUT}} & {\textbf{Vocal}} & {\textbf{Mucho}} & {\textbf{Video}} & {\textbf{Ego}} & {\textbf{NextQA}} & {\textbf{MV}} & {\textbf{MUSIC}} & {\textbf{\relp (\%)}} \\
              & {\textbf{-AQA}} & {\textbf{2017}} & {\textbf{Sound}} & {\textbf{music}} & {\textbf{MME}} & {\textbf{Schema}} & {\textbf{-MC}} & {\textbf{-Bench}} & {\textbf{-QA}} & \\
             \midrule
             Full Model &
             {\scriptsize 85.6} & {\scriptsize 71.2} & {\scriptsize 92.4} & {\scriptsize 58.9} & {\scriptsize 48.7} & {\scriptsize 49.3} & {\scriptsize 73.3} &{\scriptsize 58.4} & {\scriptsize 79.4} & {\scriptsize 100} \\
             \cmidrule{0-10}
             Shortened LLaMA &
             {\scriptsize 37.9} & {\scriptsize 12.4} & {\scriptsize 18.3} & {\scriptsize 24.6} & {\scriptsize 32.3} & {\scriptsize 8.1} & {\scriptsize 21.3} &{\scriptsize 30.0} & {\scriptsize 25.7} & {\scriptsize 34.3} \\
             ShortGPT &
             {\scriptsize \underline{64.3}} & {\scriptsize \textbf{42.3}} & {\scriptsize \textbf{77.9}} & {\scriptsize \textbf{31.6}} & {\scriptsize 43.5} & {\scriptsize \underline{31.3}} & {\scriptsize \textbf{25.0}} &{\scriptsize \textbf{45.8}} & {\scriptsize \underline{44.8}} & {\scriptsize \textbf{64.6}} \\
             OWL &
             {\scriptsize 60.0} & {\scriptsize 22.5} & {\scriptsize \underline{73.9}} & {\scriptsize 24.9} & {\scriptsize \textbf{50.0}} & {\scriptsize 27.6} & {\scriptsize 21.5} &{\scriptsize 36.8} & {\scriptsize 36.8} & {\scriptsize 55.6} \\
             ECoFLaP &
             {\scriptsize 21.3} & {\scriptsize 5.9} & {\scriptsize 22.9} & {\scriptsize 27.5} & {\scriptsize 8.7} & {\scriptsize 5.3} & {\scriptsize 19.8} &{\scriptsize 27.5} & {\scriptsize 26.8} & {\scriptsize 27.8} \\
             \cellcolor{gg}DAS (Ours) &
             \cellcolor{gg}{\scriptsize \textbf{71.0}} & \cellcolor{gg}{\scriptsize \underline{33.5}} & \cellcolor{gg}{\scriptsize 53.2} & \cellcolor{gg}{\scriptsize \underline{28.9}} & \cellcolor{gg}{\scriptsize \underline{49.3}} & \cellcolor{gg}{\scriptsize \textbf{32.2}} & \cellcolor{gg}{\scriptsize \textbf{25.0}} & \cellcolor{gg}{\scriptsize \underline{42.6}} & \cellcolor{gg}{\scriptsize \textbf{45.8}} & {\scriptsize \underline{61.3}} \\
             \bottomrule
        \end{tabular}
    }
    \caption{\small Comparison of block pruning techniques on the VideoLLaMA2 model with 25\% sparsity ratio and estimate performance on various multimodal evaluation benchmarks. The best and the second best results are in \textbf{bold} and \underline{underlined}, respectively.}
    \label{tab:structural_pruning}
\end{table*}

\paragraph{Structural Pruning.} While primary focus is on unstructured pruning of MLLMs, we also explore the applicability of our approach to structured pruning, specifically block pruning that measures the importance of blocks and removes redundant blocks accordingly. For this, we conduct experiments using the VideoLLaMA2 model. To assess block importance, we adopt several recent baselines, including Shortened LLaMA~\citep{Kim2024shortenedllama}, ShortGPT~\citep{Men2024shortgpt}, OWL, and ECoFLaP. In addition, we integrate our core component, DAS, which estimates layer or block importance based on multimodal token density distributions, into a structured pruning framework.

As shown in~\Cref{tab:structural_pruning}, DAS achieves competitive or superior performance across multiple multimodal benchmarks, demonstrating its effective extension to structured pruning scenarios. Notably, ShortGPT, which computes block importance using cosine similarity between input and output token embeddings, achieves the best overall performance. This result aligns with our approach and strengthens the insight that token distributions encode rich structural information critical for pruning. By leveraging output token density distributions, our method offers stronger generalizability and effectiveness for multimodal structured pruning, in contrast to prior unstructured pruning methods such as OWL and ECoFLAP, which rely on feature outliers or gradients for importance estimation.
\section{Conclusion}

In this paper, we investigate the critical challenges in Multimodal Large Language Model (MLLM) pruning. In our comprehensive investigations, we find that different MLLM layers have varying capabilities to encode multimodal output tokens. We also empirically observe that existing pruning methods fail to address varying modality reliance across blocks in MLLMs, resulting in suboptimal pruning outcomes. Based on these observations, we introduce \ours, a novel pruning framework that adapts both layer-wise sparsity and input activations to each layer's multimodal token attributes. We validate our method on powerful MLLMs and extensive experiments demonstrate that \ours is effective in preserving diverse multimodal abilities, even at extreme model sparsity. We believe our work offers a strong foundation for future advancements in MLLM pruning, enabling the deployment of recent MLLMs in source-constrained scenarios.
\section*{Limitations}
While \ours shows promising performance on recent MLLMs, including LLaVA-NeXT, VideoLLaMA2, and LLaVA-OneVision, this work mainly focuses on unstructured pruning for MLLMs. However, both the main body and the appendix reveal results consistent with recent structured pruning methods, indicating the potential of our core intuitions. Although we conducted the block pruning experiment, the performance of our approach was not particularly strong. Future research will investigate how our approach can be extended and improved in this direction to enhance the applicability and efficiency of MLLM pruning techniques.

Despite evaluating \ours on several MLLMs, MLLMs handling other modalities, such as point clouds, molecules, and proteins, or MLLMs incorporating Q-Former structures remain unexplored. Evaluating our approach across a border range of settings would further validate its generalizability. Additionally, our study primarily examines the performance-sparsity trade-offs without evaluating the impact on hardware efficiency. While unstructured pruning can theoretically reduce computation, future research in structured pruning should explore how \ours can deliver practical benefits in terms of latency and deployment efficiency.


\bibliography{custom}
\clearpage
\appendix
\begin{table*}[t]
    \tiny
    \centering
    \resizebox{\textwidth}{!}{
        \renewcommand{\arraystretch}{1.2}
        \renewcommand{\tabcolsep}{6.5pt}
        \begin{tabular}{l|c c c c|c c c c a}
             \toprule
              & \multicolumn{4}{c|}{\textbf{Interleaved Image}} & \multicolumn{4}{c}{\textbf{Video}} & \\
             {\textbf{Method}} & {\textbf{Muir}} & {\textbf{Mantis}} & {\textbf{BLINK}} & {\textbf{Text-rich}} & {\textbf{Video}} & {\textbf{Ego}} & {\textbf{NextQA}} & {\textbf{MV}} & {\textbf{\relp (\%)}} \\
              & {\textbf{Bench}} & {} & {} & {\textbf{VQA}} & {\textbf{MME}} & {\textbf{Schema}} & {\textbf{-MC}} & {\textbf{-Bench}} & \\
             \midrule
             Full Model &
             {\scriptsize 41.8} & {\scriptsize 64.2} & {\scriptsize 48.4} & {\scriptsize 80.1} & {\scriptsize 60.1} & {\scriptsize 56.7} & {\scriptsize 58.4} &{\scriptsize 79.4} & {\scriptsize 100} \\
             \cmidrule{0-9}
             Magnitude &
             {\scriptsize 0} & {\scriptsize 0} & {\scriptsize 0} & {\scriptsize 0} & {\scriptsize 20.7} & {\scriptsize 0} & {\scriptsize 0} & {\scriptsize 19.8} & {\scriptsize 7.4} \\
              SparseGPT &
             {\scriptsize \textbf{41.1}} & {\scriptsize 51.9} & {\scriptsize \textbf{46.2}} & {\scriptsize 63.6} & {\scriptsize \textbf{55.2}} & {\scriptsize \textbf{55.1}} & {\scriptsize 52.3} & {\scriptsize 75.2} & {\scriptsize 90.9} \\
                Wanda &
             {\scriptsize 38.6} & {\scriptsize 47.9} & {\scriptsize 44.0} & {\scriptsize \underline{65.1}} & {\scriptsize 51.5} & {\scriptsize 53.5} & {\scriptsize 49.3} & {\scriptsize 73.3} & {\scriptsize 87.0} \\
              ECoFLaP &
             {\scriptsize 40.7} & {\scriptsize \textbf{58.1}} & {\scriptsize 45.2} & {\scriptsize 64.7} & {\scriptsize \underline{54.4}} & {\scriptsize \underline{54.5}} & {\scriptsize \underline{53.2}} & {\scriptsize \textbf{76.1}} & {\scriptsize \underline{92.0}} \\
              OWL &
             {\scriptsize 35.1} & {\scriptsize 43.8} & {\scriptsize 43.5} & {\scriptsize 60.1} & {\scriptsize 50.8} & {\scriptsize 52.4} & {\scriptsize 47.6} & {\scriptsize 69.1} & {\scriptsize 82.8} \\
             \cellcolor{gg}\ours (Ours) &
             {\scriptsize \cellcolor{gg}\underline{40.9}} & {\scriptsize \cellcolor{gg}\underline{57.1}} & {\scriptsize \cellcolor{gg}\underline{45.9}} & {\scriptsize \cellcolor{gg}\textbf{69.8}} & {\scriptsize \cellcolor{gg}54.0} & {\scriptsize \cellcolor{gg}53.9} & {\scriptsize \cellcolor{gg}\underline{52.5}} & {\scriptsize \cellcolor{gg}\underline{75.2}} & {\scriptsize \textbf{92.3}} \\
             \bottomrule
        \end{tabular}
    }
    \caption{\small Comparison of pruning techniques on the LLaVA-OneVision model with 60\% sparsity ratio and estimate performance on various multimodal evaluation benchmarks. The best and the second best results are in \textbf{bold} and \underline{underlined}, respectively.}
    \label{tab:llava_onevision_eval}
\end{table*}

\section{Details of Experimental Setups \label{appendix:Implementation Details}}

\paragraph{Calibration Datasets.} Following established practices in model pruning~\citep{Sung2024wanda,Frantar2023sparsegpt,Sung2024ecoflap}, we use a random subset of 128 samples from the training datasets of the target models as calibration data. For LLaVA-NeXT, we use ShareGPT4V~\citep{Chen2024sharegpt4v} as the calibration dataset. For VideoLLaMA2, we choose MUSIC-QA~\citep{Li2022musicqa} as the calibration source as its samples consist of both video and audio modalities. For LLaVA-OneVision, we use NLVR2~\citep{Suhr2019nlvr2} as it constitutes the largest portion of its training dataset.

\paragraph{Evaluation pipeline} To ensure consistency and reproducibility, the benchmarks are assessed through LMMs-Eval framework~\citep{li2024lmms} and evaluation pipelines of the models. We follow the LMMs-Eval prompt templates provided in the official GitHub repository of the LMMs-Eval to evaluate the LLaVA-NeXT and LLaVA-OneVision models. We implement the VideoLLaMA2 architecture on the LLMs-Eval framework and evaluate the model on audio and video benchmarks.

\begin{figure*}[t]
\centering
\begin{minipage}[t]{0.46\linewidth}
    \centering
    \includegraphics[width=\textwidth]{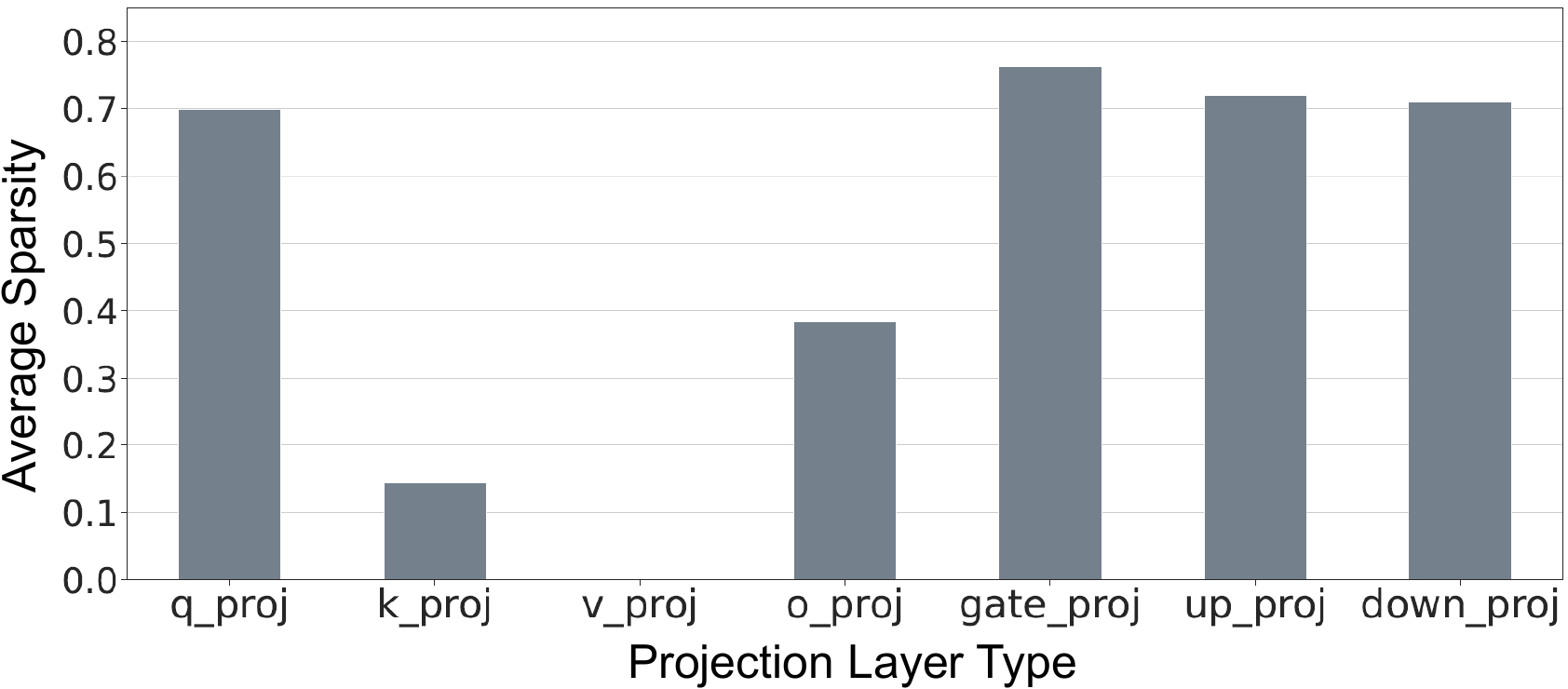}
    \par
    \caption{\small Average sparsity per projection layer type for VideoLLaMA2 at 70\% sparsity using \ours (Ours).}
    \label{fig:sub:projection_type_sparsity_analysis}
\end{minipage}
\hspace{0.1in}
\begin{minipage}[t]{0.5\linewidth}
    \centering
    \includegraphics[width=\textwidth]{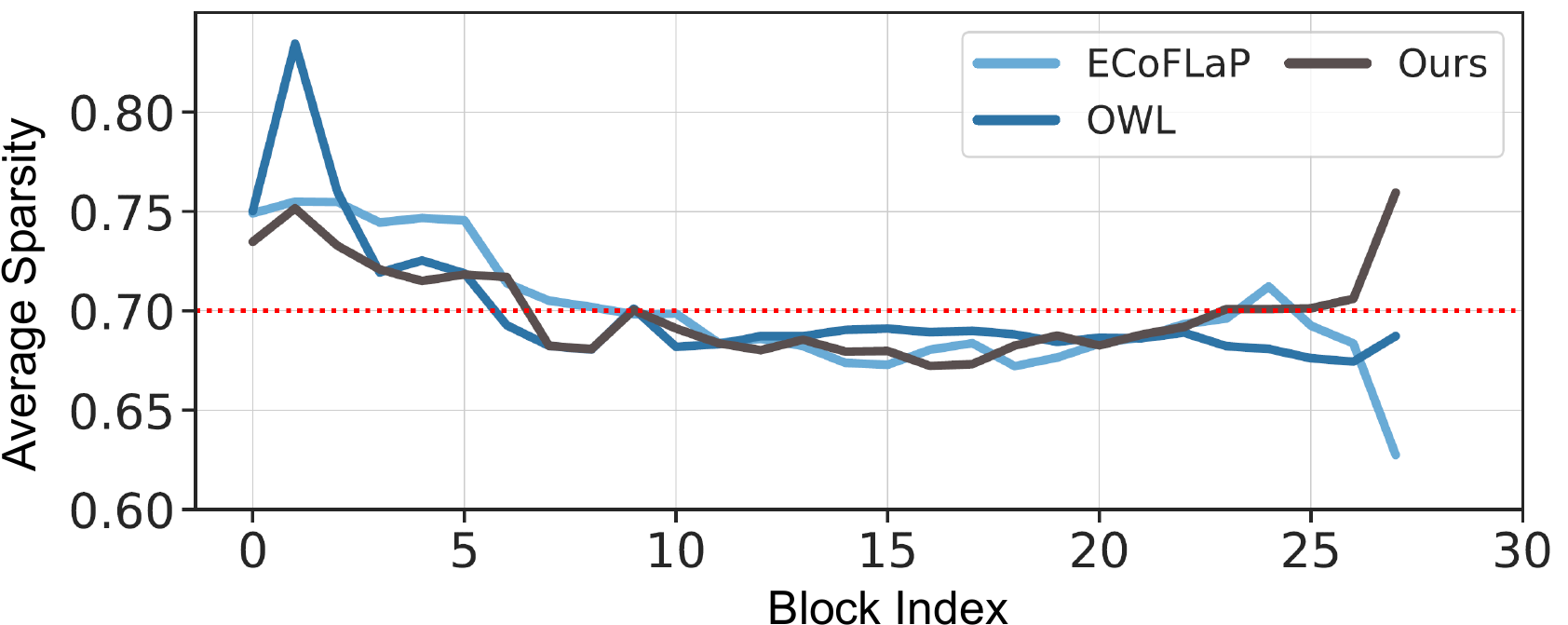}
    \par
    \caption{\small Comparison of sparsity ratio results per block for VideoLLaMA2 model at 70\% sparsity.}
    \label{fig:sub:block_wise_sparsity_analysis}
\end{minipage}
\end{figure*}
\section{Experiments on LLaVA-OneVision \label{appendix:LLaVA-Onevision_exp}}

\paragraph{Experimental Setups} We conduct additional model pruning experiments on LLaVA-OneVision~\citep{Li2024llavaonevision} with 7B parameters, which processes both interleaved images and video modalities. After pruning, we evaluate its zero-shot performance in the two modalities, following the evaluation protocols in LLaVA-OneVision: 1) interleaved images: Muirbench~\citep{Wang2024muirbench} for diverse multi-image tasks, Mantis~\citep{Jiang2024mantis} for reasoning over multiple images, BLINK~\citep{Fu2024BLINK} for multi-image visual perception tasks, and Text-rich VQA~\citep{Liu2023textrichvqa} for multi-image text recognition; 2) video: VideoMME and NeXTQA-MC for diverse video domains and durations, EgoSchema for long video understanding, and MVBench for spatio-temporal understanding.

\paragraph{Experimental Results}
\Cref{tab:llava_onevision_eval} summarizes performance of LLaVA-OneVision at a 50\% sparsity ratio. Across 6 out of 8 interleaved images understanding and video benchmarks, \ours ranks either first or second. On average, \ours surpasses the Wanda and the strongest baseline by 5.3 pp and 0.3 pp, respectively, in relative performance. These results demonstrate that the effectiveness of our method can be transferred to the pruning of other recent MLLMs with different multimodal settings, further supporting the universality of our approach.


\section{In-Depth Analysis on Layer-wise Sparsity Ratios \label{appendix:sparsity_analysis}}

\subsection{Sparsity of Projection Layer Type}

In~\Cref{fig:similarity_trend}, we observe significant variations in intra- and inter-modality diversities across different projection layer types and leverage these variations to estimate layer importance. In this ablation study, we examine the sparsity results of different projection layer types in MLLMs. 
~\Cref{fig:sub:projection_type_sparsity_analysis} presents the average sparsity ratios across blocks for each projection layer type in VideoLLaMA2 pruned at 70\% sparsity using \ours. Our analysis reveals that in the MHA module, comprising query, key, value, and output projection layers, the value projection layer consistently exhibits the lowest sparsity ratio. In contrast, in the FFN module, which consists of gate, up, and down projection layers, all projection layers exhibit relatively high sparsity levels compared to the layers in the MHA module, with the gate projection layer showing the highest value.

These findings suggest that FFN modules are more robust in pruning than MHA modules, which aligns with the recent work~\citep{Zhang2024finercut} on pruning either MHA or FFN modules in LLMs.
Moreover, our results imply that value projection layers may play a more crucial role in encoding token features compared to other projection layers, containing more critical parameters necessary for preserving the performance of MLLMs.

Interestingly, the aforementioned trend in average sparsity per layer type shown in~\Cref{fig:sub:projection_type_sparsity_analysis} is consistent with the result in EvoPress~\citep{Sieberling2024evopress}, which uses an evolutionary algorithm to find the optimal sparsity levels across LLM layers or blocks. This alignment supports our core intuition that layers producing higher multimodal output token diversity should retrain more parameters during pruning to preserve their capability to encode richer multimodal information. Moreover, \ours uncovers this trend through a simpler approach based on average cosine distances among multimodal tokens, while EvoPress requires iterative exploration in the vast sparsity ratio solution space. This indicates that output token distributions can offer an efficient and insightful basis for estimating layer importance, leading to more effective pruning outcomes.

\subsection{Block-wise Average Sparsity} To further investigate layer-wise sparsity strategies, we analyze sparsity ratios across block depths for VideoLLaMA2 at 70\% sparsity ratio, as determined by ECoFLaP, OWL, and \ours. The results illustrated in~\Cref{fig:sub:block_wise_sparsity_analysis} indicate that all three methods follow a similar sparsity trend, where the initial blocks have high sparsity and intermediate blocks exhibit moderate sparsity. Notably, \ours exhibits higher sparsity in the last blocks. 

These trends diverge from typical observations in LLM pruning. Recent studies suggest that intermediate blocks generally contain large redundancy, where pruning these blocks results in a mere impact on LLM performance. In contrast, the studies show that pruning early or final blocks leads to substantial performance degradation~\citep{Men2024shortgpt,Zhong2024blockpruner}. However, in MLLM pruning in~\Cref{fig:sub:block_wise_sparsity_analysis}, we observe an opposite pattern, particularly with \ours.

This difference can be explained by the attention distribution trends shown in~\Cref{fig:attention_trend}. Our analysis reveals that both visual and language attention scores are notably high in the intermediate blocks, indicating active multimodal interactions. Thus, these blocks would require lower sparsity to align with the increased multimodal integration occurring at this stage. In contrast, in the later blocks, only their language attention scores are high while visual attention scores are low. We hypothesize that at this stage, the MLLM primarily focuses on language generation, reducing the need for multimodal processing, which aligns with token reduction studies in MLLMs~\citep{Chen2024fastv}. This suggests that non-language modality information becomes redundant in these layers, requiring comparably fewer parameters.

These findings support the necessity of pruning strategies specifically designed for MLLMs, as their architectural and functional characteristics differ significantly from those of LLMs. Conventional LLM pruning techniques may therefore be suboptimal for multimodal models.

\section{Robustness of \ours}
\begin{table*}[t]
    \tiny
    \centering
    \resizebox{\textwidth}{!}{
        \renewcommand{\arraystretch}{1.2}
        \renewcommand{\tabcolsep}{7.5pt}
        \begin{tabular}{l c c c c c c c a}
             \toprule
             {\textbf{Method}} & {\textbf{MME-}} & {\textbf{MME-}} & {\textbf{ChartQA}} & {\textbf{AI2D}} & {\textbf{MMMU}} & {\textbf{Mathvista}} & {\textbf{MMBench}} & {\textbf{\relp (\%)}}\\
              & {\textbf{cognition}} & {\textbf{perception}} & & & & & & \\
             \midrule
             Full Model &
             {\scriptsize 376.8} & {\scriptsize 1588.3} & {\scriptsize 69.2} & {\scriptsize 71.7} & {\scriptsize 40.1} & {\scriptsize 36.2} & {\scriptsize 72.2} & {\scriptsize 100} \\
             \cmidrule{0-8}
              SparseGPT &
             {\scriptsize \underline{313.0}} & {\scriptsize \underline{1447.9}} & {\scriptsize \underline{65.4}} & {\scriptsize 64.0} & {\scriptsize 34.2} & {\scriptsize \underline{31.6}} & {\scriptsize 65.7} & {\scriptsize \underline{88.8}} \\
             Wanda &
             {\scriptsize 286.7} & {\scriptsize 1348.3} & {\scriptsize 63.9} & {\scriptsize 63.9} & {\scriptsize \underline{35.2}} & {\scriptsize 30.3} & {\scriptsize 63.8} & {\scriptsize 86.1} \\
              ECoFLaP &
             {\scriptsize 273.0} & {\scriptsize 1439.9} & {\scriptsize \textbf{65.7}} & {\scriptsize \textbf{66.0}} & {\scriptsize 35.3} & {\scriptsize 30.8} & {\scriptsize \underline{66.2}} & {\scriptsize 87.9} \\
              OWL &
             {\scriptsize 271.9} & {\scriptsize 1335.6} & {\scriptsize 61.9} & {\scriptsize 61.0} & {\scriptsize 33.6} & {\scriptsize 29.2} & {\scriptsize 61.0} & {\scriptsize 82.8} \\
             \cellcolor{gg}\ours (Ours) &
              \cellcolor{gg}{\scriptsize \textbf{337.2}} & \cellcolor{gg}{\scriptsize \textbf{1461.3}} & \cellcolor{gg}{\scriptsize 64.8} & \cellcolor{gg}{\scriptsize \underline{64.8}} & \cellcolor{gg}{\scriptsize \textbf{35.9}} & \cellcolor{gg}{\scriptsize \textbf{32.2}} & \cellcolor{gg}{\scriptsize \textbf{66.5}} & {\scriptsize \textbf{90.9}} \\
             \bottomrule
        \end{tabular}
    }
    \caption{\small Comparison of pruning techniques on the LLaVA-NeXT model with 50\% sparsity ratio. Reported values are the average over three different calibration sets. The best and the second best results are in \textbf{bold} and \underline{underlined}, respectively.}
    \label{tab:significance_test_avg}
\end{table*}

\begin{table*}[t]
    \tiny
    \centering
    \resizebox{\textwidth}{!}{
        \renewcommand{\arraystretch}{1.2}
        \renewcommand{\tabcolsep}{7.5pt}
        \begin{tabular}{l c c c c c c c a}
             \toprule
             {\textbf{Method}} & {\textbf{MME-}} & {\textbf{MME-}} & {\textbf{ChartQA}} & {\textbf{AI2D}} & {\textbf{MMMU}} & {\textbf{Mathvista}} & {\textbf{MMBench}} & {\textbf{\relp (\%)}}\\
              & {\textbf{cognition}} & {\textbf{perception}} & & & & & & \\
             \midrule
             Full Model &
             {\scriptsize -} & {\scriptsize -} & {\scriptsize -} & {\scriptsize -} & {\scriptsize -} & {\scriptsize -} & {\scriptsize -} & {\scriptsize -} \\
             \cmidrule{0-8}
              SparseGPT &
             {\scriptsize 23.1} & {\scriptsize 16.9} & {\scriptsize \textbf{0.12}} & {\scriptsize 0.79} & {\scriptsize \underline{0.75}} & {\scriptsize 1.41} & {\scriptsize 0.90} & {\scriptsize \underline{0.62}} \\
             Wanda &
             {\scriptsize 19.3} & {\scriptsize 10.6} & {\scriptsize 0.88} & {\scriptsize 0.47} & {\scriptsize 1.06} & {\scriptsize \underline{0.25}} & {\scriptsize \underline{0.23}} & {\scriptsize 0.75} \\
              ECoFLaP &
             {\scriptsize 16.5} & {\scriptsize \underline{9.24}} & {\scriptsize 0.18} & {\scriptsize \textbf{0.12}} & {\scriptsize 0.63} & {\scriptsize \textbf{0.15}} & {\scriptsize \textbf{0.09}} & {\scriptsize 0.90} \\
              OWL &
             {\scriptsize \textbf{2.5}} & {\scriptsize 27.7} & {\scriptsize 1.10} & {\scriptsize 2.61} & {\scriptsize 1.49} & {\scriptsize 1.43} & {\scriptsize 2.68} & {\scriptsize 2.66} \\
             \cellcolor{gg}\ours (Ours) &
              \cellcolor{gg}{\scriptsize \underline{4.1}} & \cellcolor{gg}{\scriptsize \textbf{8.42}} & \cellcolor{gg}{\scriptsize \textbf{0.12}} & \cellcolor{gg}{\scriptsize \underline{0.15}} & \cellcolor{gg}{\scriptsize \textbf{0.32}} & \cellcolor{gg}{\scriptsize 0.31} & \cellcolor{gg}{\scriptsize 0.35} & {\scriptsize \textbf{0.15}} \\
             \bottomrule
        \end{tabular}
    }
    \caption{\small Standard deviation of the performance of pruning techniques on the LLaVA-NeXT model with 50\% sparsity ratio. Reported values are the average over three different calibration sets. The best and the second best results are in \textbf{bold} and \underline{underlined}, respectively.}
    \label{tab:significance_test_std}
\end{table*}

In this section, we evaluate the robustness of pruning methods under variations in the calibration set. Specifically, we conduct additional experiments on the LLaVA-NeXT model with 50\% sparsity ratio using two additional randomly selected calibration sets. Including the original calibration set, as reported in~\Cref{tab:llava_next_eval}, we compute the mean and standard deviation of performance across three trials. 

As shown in~\Cref{tab:significance_test_avg}, \ours consistently outperforms all baseline methods, including SparseGPT, the second-best method, across most benchmarks, achieving the highest average relative performance. In~\Cref{tab:significance_test_std}, we observe that \ours not only achieves the highest average score but also the smallest standard deviation across most benchmarks. Notably, the conservative estimate of \ours's performance ($90.9-0.15=90.75$) is still higher than the optimistic estimate of SparseGPT's ($88.8+0.62=89.42$), suggesting a statistically significant margin that is unlikely to arise from random variations in calibration set selection.

Moreover, the consistently low standard deviations achieved by \ours suggest that it is less sensitive to calibration data variability compared to other methods. These findings provide strong evidence of both the robustness and effectiveness of our proposed approach.

\section{Visualizing Multimodal Selection \label{appendix:token_selection}}
\begin{figure*}[t]
    \centering
    \includegraphics[width=\linewidth]{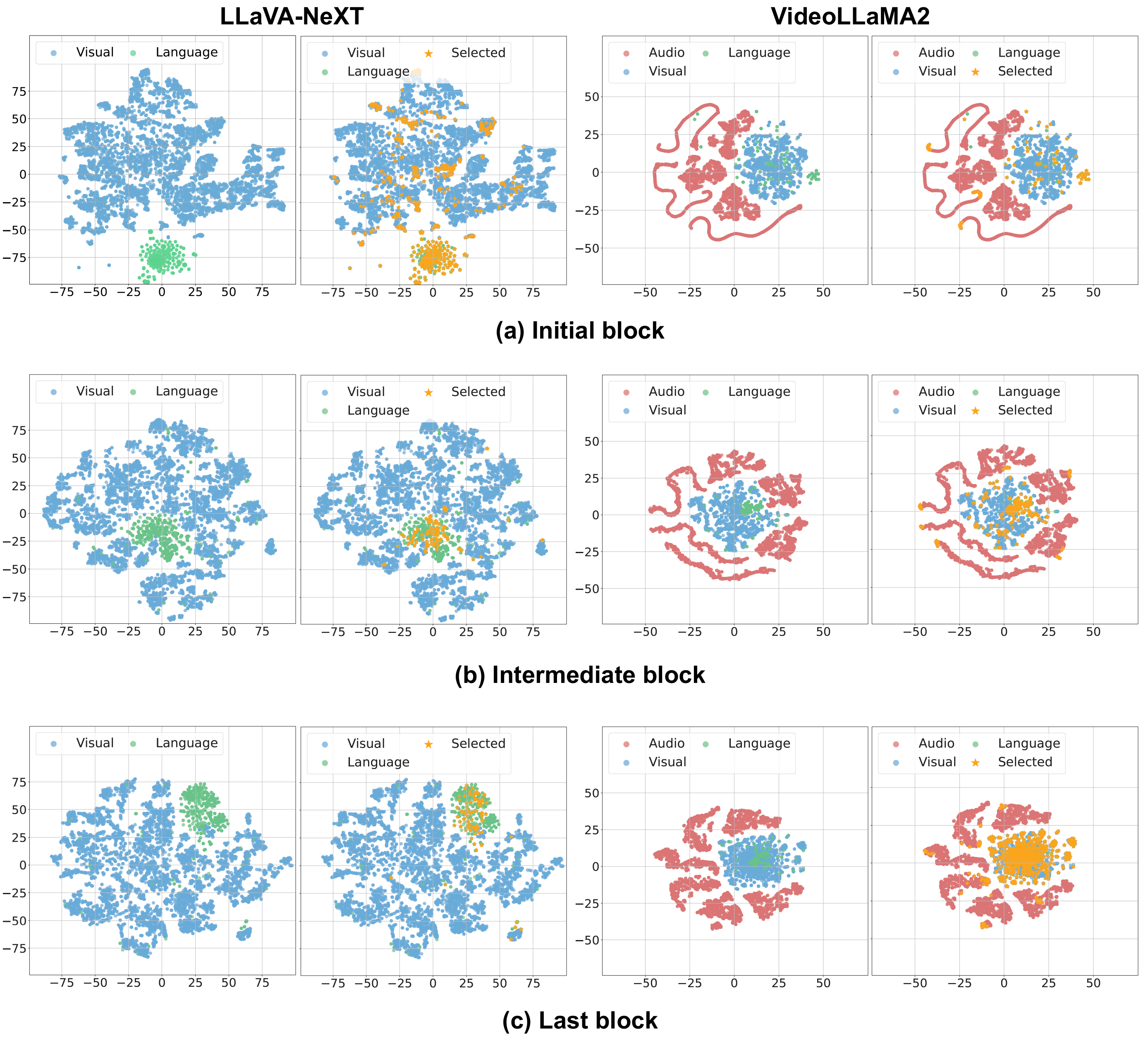}
    \par
    \vspace{-0.025in}
    \caption{\small Token selection results of Adaptive Multimodal Input Activation. We use t-SNE visualization for multimodal output token space across value projection layers of initial, intermediate, and last blocks of LLaVA-NeXT and VideoLLaMA2. 
    }
    \vspace{-0.05in}
    \label{fig:sub:token_selection}
\end{figure*}
In~\Cref{fig:sub:token_selection}, we illustrate token selection results using the AMIA selection strategy across the initial, intermediate, and last blocks of LLaVA-NeXT and VideoLLaMA2. Specifically, we visualize the multimodal output token spaces from the value projection layers in each block. Each block exhibits different modality distributions and selection results, showcasing the varying multimodal processing demands across different blocks. For example, in LLaVA-NeXT, the initial block selects both visual and language tokens, indicating its need for visual information during multimodal processing. However, in the intermediate and final blocks of the LLaVA-NeXT, comparably fewer visual tokens are selected, suggesting that these blocks assign less importance to visual information compared to language information. In VideoLLaMA2, the last block continues to rely on both language and visual tokens. We attribute this to VideoLLaMA2's architecture, which processes video inputs through spatial-temporal aggregation. This design yields more compact and informative video tokens compared to approaches that simply segment video into small image pacthes~\citep{Cheng2024videollama2,Li2024llavanextinterleave}.

\end{document}